\documentclass{article}
\usepackage{kr}

\usepackage{times}
\usepackage{soul}
\usepackage{url}
\usepackage[hidelinks]{hyperref}
\usepackage[utf8]{inputenc}
\usepackage[small]{caption}
\usepackage{graphicx}
\usepackage{amsmath}
\usepackage{amsthm}
\usepackage{booktabs}
\usepackage{algorithm}
\usepackage{algorithmic}
\begin{document}

\title{Global Mixup: Eliminating Ambiguity with Clustering}

\author{%
Xiangjin Xie$^1$\and
Yangning Li$^1$\and
Wang Chen$^3$\and
Kai Ouyang$^1$ \and
Li Jiang$^1$ \and
Haitao Zheng$^{1,2,*}$
\affiliations
$^1$Tsinghua University, China\\
$^2$Pengcheng Laboratory, 518055, China\\
$^3$Columbia University, USA\\
\emails
\{xxj20, lyn20, oyk20, jl20\}@mails.tsinghua.edu.cn,
wc2794@columbia.edu,
zheng.haitao@sz.tsinghua.edu.cn
}

\maketitle

\begin{abstract}
Data augmentation with \textbf{Mixup} has been proven an effective method to regularize the current deep neural networks. Mixup generates virtual samples and corresponding labels at once through linear interpolation. However, this one-stage generation paradigm and the use of linear interpolation have the following two defects:
(1) The label of the generated sample is directly combined from the labels of the original sample pairs without reasonable judgment, which makes the labels likely to be ambiguous.
(2) linear combination significantly limits the sampling space for generating samples.
To tackle these problems, we propose a novel and effective augmentation method based on global clustering relationships named \textbf{Global Mixup}.
Specifically, we transform the previous one-stage augmentation process into two-stage, decoupling the process of generating virtual samples from the labeling. And for the labels of the generated samples, relabeling is performed based on clustering by calculating the global relationships of the generated samples.
In addition, we are no longer limited to linear relationships but generate more reliable virtual samples in a larger sampling space.
Extensive experiments for \textbf{CNN}, \textbf{LSTM}, and \textbf{BERT} on five tasks show that Global Mixup significantly outperforms previous state-of-the-art baselines. Further experiments also demonstrate the advantage of Global Mixup in low-resource scenarios.
\end{abstract}

\begingroup
\renewcommand\thefootnote{*}
	\footnotetext{Contact Author}
\endgroup

\section{Introduction}
\label{sec:introduction}
Although deep learning systems have achieved great results, they are error-prone when lacking sufficient training data. Data augmentation can effectively alleviate this problem by generating virtual data transformed from the training sets. In Natural Language Processing (NLP), the data augmentations methods consist mainly of rule-based and deep learning-based. Rule-based methods usually rely on manually designed paradigms, such as synonym replacement \cite{zhang2015character}, random noise Injection, deletion, and insert \cite{eda}. Deep learning-based methods leverage deep models to generate new data automatically. Recently proposed Mixup \cite{mixup}, and its variants\cite{seqmix,manifold,wordmixup} further improve the efficiency and robustness of models by semantic linear interpolation, which generate more virtual samples in semantic feature space. 

While achieving promising results, Mixup still has some limitations. First, the sample generation and label determination of Mixup are integrated, and mixup only considers the linear relationship between the two original samples, which leads to ambiguity of the labels of the generated samples. Second, the distribution of the data generated by the linear relationships lacks diversity, and it can only be distributed over convex combinations consisting of existing samples, which limits the regularization ability of Mixup.
\begin{figure}
\centering
\includegraphics[angle=0, width=0.45\textwidth]{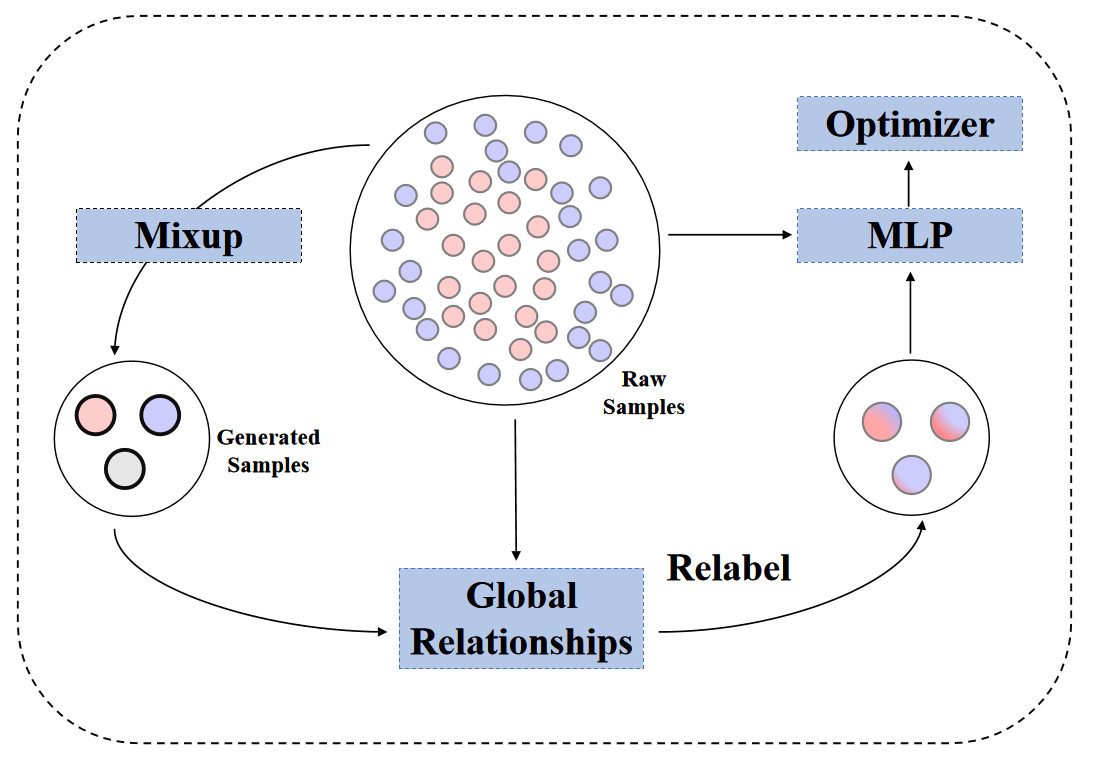} 
\caption{Illustration of the two-stage paradigm of Global Mixup, a new paradigm for data augmentation.} 
\label{fig:1} 
\end{figure}
To address these problems, we propose \textbf{Global Mixup}, a data augmentation method that takes into account the global relationships of the samples by using the clustering relationships of the samples.       
Global Mixup separates the traditional sample generation and labels determination for data augmentation into two stages. The generated samples will not directly inherit the labels of the original samples or sample pairs. Specifically, for the problem of the labels of the generated samples, Global Mixup labels the generated sample based on its global relationships with the original set of samples in the training set, encouraging the labels of the generated samples to focus on the clustering relationship with the original samples in the training set. Thus, the generated samples are uniquely labeled through global relationships, mitigating the ambiguity inherent in Mixup, which only considers local linear relationships. 
Then, for the distribution problem of the generated samples, because Global Mixup's sample generation and labeling processes are separate, the generated data of Global Mixup can be obtained from broader distributions to scale the training data more efficiently. \textbf{Figure} \ref{fig:1} shows the process of Global Mixup, it's a new paradigm for data augmentation, through split sample generation and label determination, the generated samples will get more accurate labels which can reduce the error optimization during model training.

Experiments on the classical models and pre-trained model show that Global Mixup significantly outperforms the rule-based methods and Mixup \cite{wordmixup} on different text classification tasks. The advantage of this method is more evident in low-resource scenarios, using 23\% of the training data on SST-1 and 36\% of the training data on TREC exceeds the accuracy of baseline with all training data.

In a nutshell, our main contributions are three-fold:

(1) To the best of our knowledge, we were the first to split sample generation and label determination into two separate phases in augmentation and obtain more accurate labels for the generated samples based on clustering relationships.

(2) We present a novel data augmentation approach termed Global Mixup, which implies stable labels to the virtual samples and avoids the emergence of ambiguous, overconfident labels in linear interpolation methods. Moreover, theoretically, because of the separation of the sample generation and labeling processes, Global Mixup is capable of labeling arbitrary samples, not limited to those inside convex combinations.

(3) Extensive experiments on five datasets and three models (including pre-trained models) demonstrate the effectiveness of Global Mixup, especially in few-shot scenarios.

\section{Related Work}
Data augmentation has become a prevalent research topic in recent years to solve the data scarcity problem. Automatic data augmentation has improved significant performance on various tasks such as computer vision \cite{1998transformation,zhang2015character} and speech tasks \cite{speech}. However, only rare research exploits data augmentation in natural language processing tasks because of the high complexity of language and words' discreteness. Dominant data augmentation and Interpolation-based data augmentation are two main kinds of methods that can be introduced into NLP tasks.

\subsection{Dominant data augmentation}

Dominant data augmentation focuses on generating new sentences similar to the labeled data by introducing external knowledge:

\subsubsection{Rule-based data augmentation}
Rule-based methods generate samples by transforming the original sample with human-designed rules, such as \cite{eda} using synonym substitution, random insertion, random exchange, and random deletion. \cite{char} replace words based on an English thesaurus. \cite{coulombe2018text} proposes synonymous substitution, according to the types of words suitable for replacement: adverbs, adjectives, nouns, verbs, and simple pattern matching conversion and grammar tree conversion using regular expressions to generate new sentences. Other works \cite{word2vecaug} also try to use the most similar words for text replacement based on pre-trained word vectors such as Glove\cite{pennington2014glove}, Word2vec\cite{word2vec}. 
\subsubsection{Generation-based data augmentation}
Generation-based methods focus on generating sentences based on language models. \cite{sennrich2016improving} utilize an automatic back-translation to pair monolingual training data as additional parallel training data. \cite{gan} use generative adversarial networks (GANs) \cite{ganmo} to generate new training examples from existing ones \cite{gan}. \cite{qanet} consider back-translation based on a neural machine translation model. \cite{xiedata} introduces data noise in neural network language models. Recently, pre-trained language models are also used to generate new labeled data based on contextual information \cite{contextual}. \cite{conditional} apply the conditional BERT \cite{bert} model to enhance contextual augmentation. However, the data generated by dominant data augmentation methods are similar to the original data, leading to the model still learning similar patterns. Therefore, the model cannot handle data scarcity problems when the test data distribution differs from the training data. To address this problem, Global Mixup extends the training data to different distributions based on interpolation-based data augmentation.

\subsection{Interpolation-based data augmentation}

Interpolation-based data augmentation has been proposed in Mixup \cite{mixup}. Mixup extends the training data by training a neural network on convex combinations of pairs of examples and their labels, as shown in \ref{sssec:num1}. Mixup has achieved relative success in many computer vision tasks. 
Recently, more researches have focused on utilizing Mixup to improve the model's performance in NLP tasks. Mixup variants \cite{manifold,summers} use interpolation in the hidden representation to capture higher-level information and obtain smoother decision boundaries. wordMixup \cite{wordmixup} performs interpolation on word embeddings and sentence embeddings. SeqMix \cite{seqmix} generates subsequences along with their labels by using linear interpolation. These methods optimize Mixup by modifying the data generation based on Mixup and proven effective. However, linear interpolation methods only take the relationships between two samples for the labels. To address this problem, Global Mixup introduces clustering relationships into data augmentation to assign labels to the generated samples. What is more, Global Mixup has two stages to generate samples and labels respectively so that the other methods, such as Mixup and its variants, can improve the quality of their labels based on Global Mixup.

\section{Methodology}

\begin{figure*}
\centering

\includegraphics[angle=0, width=1\textwidth]{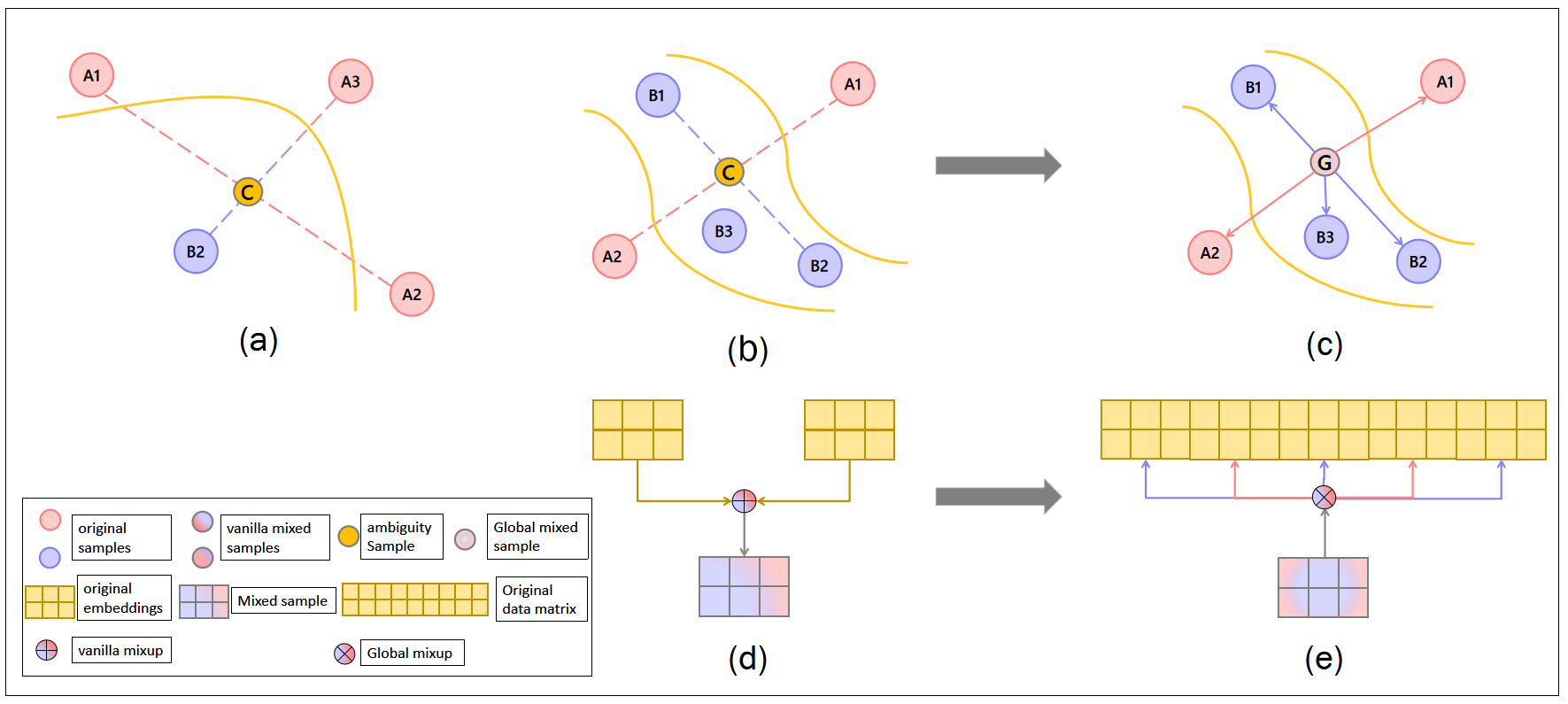} 

\caption{An overview of Global Mixup: (a) and (b) represent ambiguous cases in Vanilla Mixup, and (c) represents the Global Mixup with relabeling of the ambiguous samples in (b). (d) represents the way (a) and (b) generate samples and labels, and (e) represents the way (c) relabels the samples.} 
\label{fig:2} 
\end{figure*}

We present overviews of the method composition of Global Mixup in \textbf{Figure}~\ref{fig:2}. 
The purpose of Global Mixup is to separate the sample generation and label determination process of data augmentation and to obtain accurate sample labels by sample similarity, and encouraging the model to focus on the clustering relationships of samples to resolve the ambiguity of linear interpolation. To achieve this, we inherit the way Mixup generates virtual samples and optimize the way it labels samples.


\subsection{Preliminaries} \label{sssec:num1}

We first briefly describe the original Mixup \cite{mixup} and the variant of Mixup for text classification, wordMixup \cite{wordmixup}.

\textbf{Mixup} \cite{mixup} is the first data augmentation method proposed for image classification tasks that implements linear interpolations to generate virtual samples to encourage models to behave linearly in-between training examples. 
Let $(x, y)$ denote a sample of training data, where $x$ is the raw input samples and $y$ represents the one-hot label of $x$.
In short, the Mixup generates virtual training samples $(\tilde{x}, \tilde{y})$ can be formulated as follows:
\begin{equation}
\begin{split}
& \tilde{x}=\lambda x_{i}+(1-\lambda) x_{j}, \\
& \tilde{y}=\lambda y_{i}+(1-\lambda) y_{j}, \\
\end{split}
\end{equation}
where $(x_i, y_i)$ and $(x_j, y_j)$ are two original samples drawn at random from training data, the mixing coefficient $\lambda \sim Beta (\alpha, \alpha)$, for $\alpha \in(0, \infty)$, and $Beta$ means the Beta distribution. Unlike the original sample, which uses hard labels, the generated virtual data uses soft labels. Then both the generated virtual and the original data are used to train the network.

\textbf{wordMixup} \cite{wordmixup} is a linear Mixup method for text classification.
Firstly, it converts all sentences into embedding matrix and pads them to the same length. 
For a set of training texts, they will all be represented as the same dimensional matrix $B \in R^{N \times d}$, where $N$ represents the length of each text after padding and $d$ represents the dimension of the vector for each word. 
Secondly, $(B_i, y_i)$ and $(B_j, y_j)$ are drawn at random from original train set, where $y_i$ and $y_j$ denote the corresponding class label of the sentence using one-hot representation. 
In short, the wordMixup generates virtual training samples $(\widetilde{B},\widetilde{y})$ can be formulated as follows:
\begin{equation}
\begin{split}
&\widetilde{B}=\lambda B_{i}+(1-\lambda) B_{j}, \\
&\widetilde{y}=\lambda y_{i}+(1-\lambda) y_{j}, \\
\end{split}
\label{wordMixup}
\end{equation}
where the mixing coefficient $\lambda \sim Beta (\alpha, \alpha)$ is the same as in the Mixup, and $\alpha$ is set as 1. 

\subsection{Global Mixup}\label{GLM}


In \textbf{Vanilla Mixup}, including Mixup and the variations of Mixup, the generated virtual samples may have label ambiguity problems in the regions where linear interpolation of randomly selected original samples are intersections.
For example, the Mixup aims to generate a virtual sample by linear interpolation as shown in \textbf{Figure} \ref{fig:2}, but the same virtual sample which comes from different pairs of original samples will receive different labels as shown in the \textbf{Figure} \ref{fig:2} (a) and (b). 
When selecting extremely distinct sample pairs mixup and intersection occurs, the generated virtual samples are similar, but the labels are opposite and overconfident. 
We call this phenomenon that the label gap between similar virtual samples generated based on different sample pairs is too large, the label ambiguity problem. 

To tackle the label ambiguity problem, we propose to calculate the global relationships of the generated virtual samples.
Specifically, as shown in \textbf{Figure} \ref{fig:2} (b), When we generate the same virtual sample $C$ based on two sample pairs $(A_1, A_2)$ and $(B_1, B_2)$ that have different labels, if we are using \textbf{Vanilla Mixup}, there will be a conflict in labeling the virtual sample $C$ because the sample pair $(A_1, A_2)$ corresponds to a different label than $(B_1, B_2)$.
Moreover, when the distribution of generated virtual samples is similar, ambiguity phenomenon often occurs.
As shown in \textbf{Figure} \ref{fig:2}(c), using \textbf{Global Mixup} for the generated virtual sample $G$ will calculate the global relationships of $G$ with all original training samples to generate the labels, so it will get a globally unique label, so that eliminates ambiguity. 
Also, labeling and generation are independent when using Global Mixup. 
The generated samples can not be limited to the linear relationships of the original samples, which provides more options for generated samples in the distribution space.

Specifically, training the neural networks using Global Mixup mainly consists of the following four steps:

\textbf{Raw Samples Selection}: In this step, we randomly select a part of the sample pairs $(B_i, y_i)$ and $(B_j, y_j)$ from training data as raw materials for making virtual samples. 

\textbf{Raw Mixed Samples Generation}: After randomly selecting the raw samples, we perform linear interpolation on them and generate virtual training samples$(\tilde{B}, \tilde{y}) $ as shown in Equation \ref{wordMixup}. For simplicity, the Vanilla Mixup samples generation method is used here.

\textbf{Labels Reconfiguration}: In this part, we select a part of raw mixed sample for relabeling, usually choosing those raw mixed samples with overconfident labels. 
Specifically, we select samples with label $\tilde{y}$ satisfying $\operatorname{{\arg \max }} \tilde{y} \geq \theta$ from the generated virtual sample set for relabeling, which means that the labels of overconfident virtual samples will be recalculated. The selection parameter $\theta \in [1/c, 1]$, $c$ is the number of target label. When $\theta =1/c$, all raw mixed samples will be selected for relabeling. For example, when $\theta=1$, it reduces to the Vanilla Mixup principle. Reconstruction of the labels of these virtual samples is as follows:
\begin{equation}
y^{\star}= \sum^{T}_{t=1} P(B_t\mid  D( B_t, B'))y_t , \label{ys}
\end{equation}
where $y^{\star}$ is the new label for $B'$. $P(B_t \mid D(B_t, B'))$ is the weight of $y_t$ to generate $y^{\star}$, and $D(B_t, B')$ is the equation for computing the relationships between the training samples $B_t$ and the generated virtual sample $B'$. It can be formalized as follows:

\begin{equation}
\small
\setlength\abovedisplayskip{0cm}
\setlength\belowdisplayskip{0.2cm}
P(B_t\mid D (B_t, B'))=\frac{\exp (D( B_t, B')) }{ \sum^{T}_{i=1} \exp ({ D( B', B_i)})}, 
\label{Pd}
\end{equation}
where $T$ is the total number of samples used to calculate the global relationships, the largest $top-s$ from all computed $D$ will be used for the computation of $P$, and $P$ of $B_w(w \in {T-s} )$ will be set to 0. $s \in [2, N]$ is the number of samples referenced to calculate the global relation of $B'$. When $s=2$, only the relationships between $B'$ and the samples that generate it will be calculated. when $s=N$, all training samples are calculated, in general, We choose $T$ equal to the number of batch size.

\begin{equation}
\small
D( B_t, B')=\frac{\gamma \sum^N_{i=1}\sum^d_{j=1} B_t \cdot B'}{ \sum^N_{i=1}\sum^d_{j=1} \sqrt{B_t \cdot B_t}*\sqrt{B'\cdot B'}+\epsilon}, 
\end{equation}
where $D$ means the similarity of the matrices $B_t$ and $B'$. And $D$ can be interpreted as flattening the word embedding matrices $B_t $ and $B'$ and computing the cosine similarity of the two flattened matrices. Where $d, N $ are the dimensional parameters of the matrix $B$.  $\gamma$ is the parameter of relationships correction, and $\epsilon$ is the parameter to prevent the denominator from being 0. For the \textbf{BERT} model, due to the attention mask mechanism, we change the formula for calculating $D$ as follows, $ A \in 1*N$ represents the attention mask vector for each sentence:

\begin{equation}
\small
D( B_t, B')=\frac{\gamma  A_tB_t(A'B')^T}{ A_tB_t(A_tB_t)^T *A'B'(A'B')^T +\epsilon }.
\end{equation}

\textbf{Network Training}: Finally, we use the original samples $(B, y)$, virtual samples $(\tilde{B}, \tilde{y})$ generated by vanilla mixup and $(B', y^{\star})$ generated by Global Mixup to train the network, compute the loss value and gradients update the parameters of the neural networks.





Mathematically, Global Mixup minimizes the average of the loss function L, The loss is combined three parts of loss:
\begin{equation}
\small
\begin{split}
\mathrm{L}(\mathrm{f})=&\delta \ell_{\text {orig}}+\tau \ell_{\text {vanilla }}+\eta \ell_{\text {global }} \\
=& \underset{(B, y) \sim P}{{E}} \;\underset{(\tilde{B}, \tilde{y}) \sim P }{{E}} \;\underset{(B', y^{\star}) \sim P}{{E}} \;\underset{\lambda \sim \operatorname{Beta}(\alpha, \alpha)}{{E}} \\
&\ell(f_{k}(\operatorname{Mix}_{\lambda}(B, \tilde{B}, B'), \operatorname{Mix}_{\lambda}(y, \tilde{y}, y^{\star}))),
\end{split}
\end{equation}
where $ {(B, y) \sim P} $ represents the original distribution of the training data; $(\tilde{B}, \tilde{y}) \sim P 
$ represents the original distribution of the virtual data which generated by Vanilla Mixup; $  (B', y^{\star}) \sim P $ represents the original distribution of the virtual data which generated by Global Mixup. $P$ is the same as mixup \cite{mixup}, represents the distribution of samples. $\ell$ represents the loss function. $\lambda \sim \operatorname{Beta}(\alpha, \alpha) $ represent the distribution of $\lambda$. $\delta$, $ \tau$, $ \eta $ are discount factors and $ {f_k} $ is the network to be trained.

\subsection{Advantages of computing global relationships} 
Global Mixup policy is not limited to generating and labeling samples with actual word and sentence counterparts, so it can be unaffected by the discrete distribution of real samples. It uses two label strategies simultaneously to encourage the model to obtain more stable boundaries, and no sample generated gets an overconfident or ambiguous label. In addition, Global Mixup separates sample and label generation into two stages and relabeling by computing global relationships, so it can label any sample without being limited to convex combinations composed of training sets as in linear interpolation, this allows Global Mixup to be used with other generation methods and assists them with labeling.

\section{Experiments}
We conduct experiments on five tasks and three networks architectures to evaluate the effectiveness of Global Mixup.

\subsection{Datasets}
We conduct experiments on five benchmark text classification tasks:
\begin{enumerate}
    \item \textbf{YELP} \cite{yelp}, which is a subset of Yelp's businesses, reviews, and user data. 
    \item \textbf{SUBJ} \cite{subj}, which aims to classify the sentences as subjectivity or objectivity.
    \item \textbf{TREC} \cite{trec}, is a question dataset with the aim of categorizing a question into six question types.
    \item \textbf{SST-1} \cite{sst-2}, is Stanford Sentiment Treebank, five categories of very positive, positive, neutral, negative, and very negative, Data comes from movie reviews and emotional annotations.
    \item \textbf{SST-2} \cite{sst-2}, is the same as SST-1 but with neutral reviews removed and binary labels, Data comes from movie reviews and emotional annotations.

\end{enumerate}

\textit{Data Split}. We randomly select a subset of training data with $N =\{500, 2000, 5000\}$ to investigate the performance in few-sample scenario of Global Mixup.
Table \ref{table1} summarizes the statistical characteristics of the five data sets.

\begin{table}[htbp]
\small
	\centering  
	\begin{tabular}{|c|c|c|c|c|}  
		\hline  
		Data&c&N&V&T \\  
		\hline
		YELP&2&560000&W&38000 \\
		\hline
	
		SST-1&5&8544&1101&2210 \\
		\hline
		SST-2&2&6920&872&1821 \\
		\hline
		TREC&6&5452&W&500 \\
		\hline

		SUBJ&2&8500&500&1000 \\
		\hline

	\end{tabular}
		\caption{Summary for the datasets c: number of target labels. N: number of samples. V: valid set size. T: test set size (W
means no standard valid split was provided).}
\label{table1}
\end{table}

\subsection{Baselines and Settings}
We compare the proposed method with baselines: the original \textbf{CNNsen} \cite{Kim2014CNN}, the original \textbf{LSTMsen} \cite{lstm}, the original \textbf{BERT} \cite{bert}. And two recent augmentation methods including Easy Data Augmentation(EDA) \cite{eda}, wordMixup \cite{wordmixup}. CNNsen is a convolutional neural network and is widely used for text classification. LSTMsen is a type of the most popular recurrent neural network for natural language tasks. BERT is the most representative pre-training model in recent years. EDA is a simple but effective rule-based data augmentation framework for text. For a given sentence in training set, EDA \cite{eda}randomly choose and perform one of synonym replacements, random insertion, random swap, random deletion. wordMxiup \cite{wordmixup} is the straightforward application of linear interpolation on the word embedding layer. The model parameters are designed consistently to keep comparisons fair, and for comparative data augmentation methods, the best parameters from the extracted papers are used.

\subsection{Implementation Details}

All models are implemented with Pytorch \cite{paszke2019pytorch} and Python 3.7. We set the maximum sequence length as 256 to pad the varying-length sequences. For the parameters of Global Mixup. For the $\lambda \sim Beta(\alpha, \alpha)$ parameters, we tune the $\alpha$ from $\{0.5, 1, 2, 4, 8\}$. And to demonstrate the effectiveness of Global Mixup on a larger space, we extend $\lambda\in[-0.3, 1.3]$ with uniform distribution. We set the number of samples generated per training sample pair $T$ from$\{2, 4, 8, 16, 20, 32, 64\}$ and chose $T= 8$ for the best performance. The batch size is chosen from$\{32, 50, 64, 128, 256, 500\}$and the learning rate from$\{1e-3, 1e-4, 4e-4, 2e-5\}$. For the hyperparameter setting, we set $\theta$ from$\{1/c, 0.5, 0.6, 0.8, 0.9, 1\}$, $\gamma$ from $\{1, 2, 4, 6\}$, $\tau$ and $\eta$ from $\{1/T, 1\}$, $\epsilon =1e-5$, $\delta =1$. For the reinforced selector, we use Adam optimizer \cite{adam} for CNN and LSTM, AdamW \cite{adamw} for BERT. The pre-trained word embeddings for CNN and LSTM are 300-dimensional Glove \cite{pennington2014glove}. The parameters of \cite{bert} are derived from 'bert-base-uncased'.
For each dataset, we run experiments 10 times to report the mean and the standard deviation of accuracy (\%)

\begin{figure}
\centering
\includegraphics[angle=0, width=0.43\textwidth]{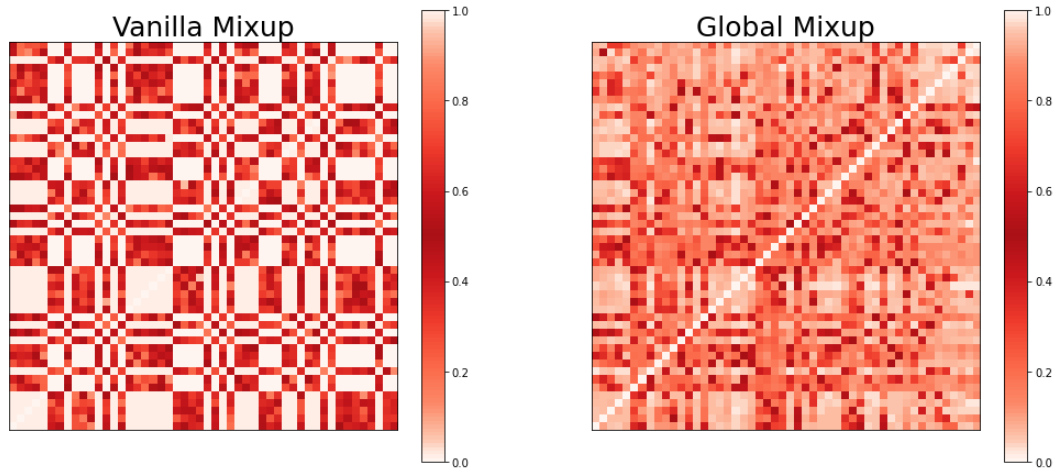} 
\caption{\small Heat map visualization of Global Mixup's relabeling. The coordinate axes represent the serial numbers of the original samples. The color represents the extreme degree of the label, the lighter the color, the more extreme the sample label. The original samples are randomly selected from YELP.} 
\label{fig:3} 
\end{figure}


\begin{table*}[htbp]
	\centering
	\begin{tabular}{cc|c|c|c|c}
		\toprule  
		Method&YELP&SST-1&SST-2&TREC&SUBJ \\ 
		\cmidrule(r){2-6} 
		CNNsen                &$92.1 \pm 0.44$ &$35.3 \pm 1.321 $& $78.5 \pm 0.56$& $95.4 \pm 0.96$ &$ 90.1 \pm 0.43$\\
		\cmidrule(r){2-6}   
		EDA                  &$92.1 \pm 0.24$  &$34.1 \pm 0.89$ & $79.3 \pm 0.49$ &$97.4 \pm 0.25$ &$91.9 \pm 0.21$\\
        \cmidrule(r){2-6}  
		wordMixup                  &$92.5\pm 0.22 $  & $36.5 \pm 0.45 $& $78.6 \pm 0.36 $ &$97.8 \pm 0.32 $ & $91.4 \pm 0.56 $\\
		\midrule  
		Global Mixup                   &$ \mathbf{93.4 \pm 0.13}$ &$\mathbf{38.5 \pm 0.20}$ & $\mathbf{80.1 \pm 0.23}$&$\mathbf{98.2\pm 0.19}$ &$ \mathbf{92.8 \pm 0.23}$\\
		\bottomrule  
	\end{tabular}
	\caption{Results on five text classification tasks for \textbf{CNN}. }
		\label{table2}
\end{table*}

\begin{table*}[htbp]
	\centering
	\begin{tabular}{cc|c|c|c|c}
		\toprule  
		Method&YELP&SST-1&SST-2&TREC&SUBJ \\ 
		\cmidrule(r){2-6} 
		LSTMsen                &$92.1 \pm 0.31$&$36.7 \pm 1.42$ &$ 79.7 \pm 0.64 $&$ 95.2 \pm 1.55 $ &$91.8\pm 0. 92 $ \\
		\cmidrule(r){2-6}   
		+EDA                         &$91.9 \pm 0.45$ &$37.8 \pm 1.33$&$81.1 \pm 0.66 $&$97.6 \pm 0.95 $&$92.1 \pm 0. 50$\\
        \cmidrule(r){2-6}  
		+wordMixup     &$ 92.8 \pm 0. 22$   &$38.4 \pm 0. 74$   &$80.6\pm 0. 42$  & $98.1 \pm 0. 63$ &$92.6 \pm 0. 42$\\
		\midrule  
		+Global Mixup          &$\mathbf{94.0 \pm 0.16}$ & $\mathbf{39.9 \pm 0.46}$ & $\mathbf{81.6 \pm 0.31} $& $\mathbf{98.6 \pm 0.35}$&$\mathbf{93.1 \pm 0.39} $\\
		\bottomrule  
	\end{tabular}
	\caption{Results on five text classification tasks for \textbf{LSTM}.}
		\label{LSTmtable}
\end{table*}

\begin{table*}[htbp]
	\centering
	\begin{tabular}{cc|c|c|c|c}
		\toprule  
		Method&YELP&SST-1&SST-2&TREC&SUBJ \\ 
		\cmidrule(r){2-6} 
		BERT                &$96.9 \pm 0.23 $&$51.9 \pm 0.92$ &$ 91.0 \pm 1.16 $&$ 99.0 \pm 0.55 $ &$96.9\pm 0.30 $ \\
		\cmidrule(r){2-6}   
		+EDA              &$97.0 \pm 0.20$ &$51.7 \pm 0.46 $&$91.3 \pm 0.55$&$98.5 \pm 0.44 $&$96.8 \pm 0. 36$\\
        \cmidrule(r){2-6}  
		+wordMixup     &$ 97.0 \pm 0.13 $   &$52.0 \pm 0. 64$   &$91.2\pm 0.56$  & $99.0 \pm 0. 16$ &$97.3 \pm 0.32$\\
		\midrule  
		+Global Mixup          &$\mathbf{97.1 \pm 0.15 }$ & $\mathbf{52.8 \pm 0.32}$ & $\mathbf{91.8\pm 0.34} $& $\mathbf{99.2 \pm 0.13}$&$\mathbf{97.5 \pm 0.35} $\\
		\bottomrule  
	\end{tabular}
	\caption{Results on five text classification tasks for \textbf{BERT}.}
		\label{BERTtable}
\end{table*}

\begin{table*}[htbp]
	\centering

	\begin{tabular}{cc|c|c|c|c}
		\toprule  
		SIZE&YELP&SST-1&SST-2&TREC&SUBJ \\ 
		\cmidrule(r){2-6} 
		500                       &$74.0 \pm 0.28 $&$27.6 \pm 1.11 $&$67.8 \pm 0.53 $&$89.8 \pm 1.87$&$83.6 \pm 0.72 $\\\specialrule{0em}{1pt}{1pt}
		+Global Mixup             &$\mathbf{81.3 \pm 0.15} $& $\mathbf{33.8 \pm 0.56}$& $69.7 \pm 0.42 $&$\mathbf{94.2 \pm 0.76}$ & $86.1 \pm0.66$\\ 
		\cmidrule(r){2-6}   
		2000                       & $  80.1 \pm 0.32$ & $30.8 \pm0.75$ & $75.5 \pm 0.42$ & $93.8 \pm 1.04 $ & $ 87.4 \pm 0.45$  \\ \specialrule{0em}{1pt}{1pt}
		+Global Mixup              & $ 85.6 \pm 0.17 $ &$35.8 \pm 0.62  $ & $ 77.4 \pm 0.61 $  &$ 96.9 \pm 0.51 $ &$ 89.4 \pm 0.31 $\\
		\cmidrule(r){2-6}  
		5000                 & $85.7 \pm 0.14$&$33.6 \pm 0.78 $&$ 77.6 \pm 0.16$ & $95.4 \pm 0.96$ &$88.7 \pm 0.41 $\\\specialrule{0em}{1pt}{1pt}
		+Global Mixup        &  $ 87.5 \pm0.13 $   &  $36.6 \pm 0.54  $  &  $79.0 \pm 0.25  $ &  $ 98.2 \pm 0.19 $  & $90.1 \pm 0.32 $\\
		\bottomrule  
	\end{tabular}
	\caption{\small Average performances (\%) across five text classification tasks with different data sizes for \textbf{CNN} with and without Global Mixup on different training set sizes. }
		\label{CNNfewtable}
\end{table*}

\begin{table*}[htbp]\vspace{-0.4cm}
	\centering
	\begin{tabular}{cc|c|c|c|c}
		\toprule  
		SIZE&YELP&SST-1&SST-2&TREC&SUBJ \\ 
		\cmidrule(r){2-6} 
		500                                     &$76.0 \pm 0.35 $&$27.3 \pm 1.26$&$68.8 \pm 1.12$&$ 88.6 \pm 1.65 $&$ 83.4\pm 1.20$\\\specialrule{0em}{1pt}{1pt}
		+Global Mixup                           &$\mathbf{82.1 \pm 0.23}$ &$30.0 \pm 0.54$&$ 69.4 \pm 0.59$  &$ 90.3 \pm 1. 44$  & $84.7 \pm 1.03 $\\ 
		\cmidrule(r){2-6}   
		2000                          & $83.1 \pm 0.31$&$ 35.1 \pm 1.17 $ &$ 76.1 \pm 0.98 $ &$92.5\pm 1.32 $ & $88.1 \pm 1.16 $\\\specialrule{0em}{1pt}{1pt}
		+Global Mixup                      & $87.2 \pm0.11$ &$36.7 \pm 0.56$&$ 77.8 \pm 0.72$  & $ 95.5 \pm 0.62 $& $89.6\pm 0.82$   \\
		\cmidrule(r){2-6}  
		5000                      &$86.2 \pm 0.23$ &$37.7 \pm 1.11 $&$ 79.1 \pm 0.73$ &$ 95.2 \pm 1.55 $ & $90.2 \pm 0.79 $\\\specialrule{0em}{1pt}{1pt}
		+Global Mixup                            & $87.8 \pm 0.21$&$38.5 \pm 0.62$& $ 79.6 \pm 0.43$&$98.6 \pm 0.35$  & $91.5 \pm 0.25$   \\
		\bottomrule  
	\end{tabular}
	\caption{\small Average performances (\%) across five text classification tasks with different data sizes for \textbf{LSTM} with and without Global Mixup on different training set sizes. }
		\label{few lstmtable}
\end{table*}

\begin{table*}[htbp]\vspace{-0.4cm}

	\centering
	\begin{tabular}{cc|c|c|c|c}
		\toprule  
		SIZE&YELP&SST-1&SST-2&TREC&SUBJ \\ 
		\cmidrule(r){2-6} 
		500                                     &$89.5 \pm 0.88 $&$35.3 \pm 2.26$&$86.4 \pm 1.42$&$92.6 \pm0.95 $&$ 94.4\pm 0.74$\\\specialrule{0em}{1pt}{1pt}
		+Global Mixup                           &$91.7 \pm 0.42$ &$\mathbf{43.0 \pm 1.15}$&$ 88.0 \pm 0.96$  &$ 97.7 \pm 0.66$  & $95.0 \pm 0.56 $\\ 
		\cmidrule(r){2-6}   
		2000                          & $92.7 \pm 0.68$&$ 47.8 \pm 1.55 $ &$89.2 \pm 1.16 $ &$98.2\pm 0.50 $ & $95.9 \pm 0.52 $\\\specialrule{0em}{1pt}{1pt}
		+Global Mixup                      & $93.2 \pm0.42$ &$49.0 \pm 0.72$&$ 89.6 \pm 0.83$  & $ 98.5 \pm 0.52 $& $96.1\pm 0.36$   \\
		\cmidrule(r){2-6}  
		5000                      &$93.4 \pm 0.55$ &$51.6 \pm 0.86 $&$ 90.5\pm 0.63$ &$ 99.0 \pm 0.55 $ & $96.2 \pm 0.42 $\\\specialrule{0em}{1pt}{1pt}
		+Global Mixup               & $94.1\pm 0.31$&$51.8 \pm 0.55$& $ 91.2 \pm 0.23$&$99.2 \pm 0.13$  & $96.8 \pm 0.45$   \\
		\bottomrule  
	\end{tabular}
	\caption{\small Average performances (\%) across five text classification tasks with different data sizes for \textbf{BERT} with and without Global Mixup on different training set sizes.}
		\label{few BERT}
\end{table*}

\subsection{Main experiment }
To demonstrate the effect of Global Mixup, we completed the main experiment on five datasets. The main results for each dataset using CNN are shown in \textbf{Table} \ref{table2}, the main results for each dataset using LSTM are shown in \textbf{Table} \ref{LSTmtable}, and the main results for each dataset using BERT are shown in \textbf{Table} \ref{BERTtable}. From the result, it is clear that our method achieves the best performance on all five data sets. For instance, compared to CNNsen, the average accuracy improvement is 3.2\% on the SST dataset Global Mixup and 3.0\% on the TREC dataset. We also observe that the Global Mixup smaller standard deviation, which validates that Global Mixup produces more stable classification boundaries. In summary, the results show a significant improvement of Global Mixup over other methods. It not only outperforms EDA, which generates real samples with rule-based, but also outperforms wordMixup, which constructs linear relationships between samples based on linear interpolation. Also, in the following ablation experiments, we show the powerful effect of Global Mixup in a few-sample scenario.

\quad In addition, to display the effect of Global Mixup relabeling, the label changes after using Global mix are shown in the \textbf{Figure} \ref{fig:3}. As shown in the figure, When the same sample pair is used to generate the same virtual sample, Vanilla Mixup and Global Mixup show dramatic differences, Vanilla Mixup generates a large number of extreme labels for the generated samples, whereas Global Mixup does not generate such a large number of extreme samples. We also observed that Global Mixup mainly relabelled the overconfident labels from Vanilla Mixup, and corrections were also made for a small number of samples that were not overly confident, by relabeling, samples sometimes even obtaining labels with the opposite polarity to that labeled by Vanilla Mixup.


\begin{figure}\vspace{-0.48cm}
\centering
\includegraphics[angle=0, width=0.43\textwidth]{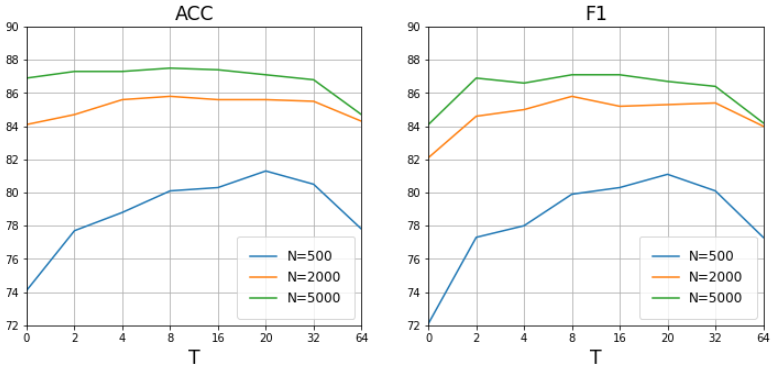}
\caption{\small Ablation study on different sample size. $N$ is the size of the data set used and $ \alpha =4$. $T=0$ means not use Global Mixup.} 
\label{fig:4} 
\end{figure}

\begin{figure}\vspace{-0.4cm}
\setlength{\belowcaptionskip}{-0.4cm} 
\centering
\includegraphics[angle=0, width=0.45\textwidth]{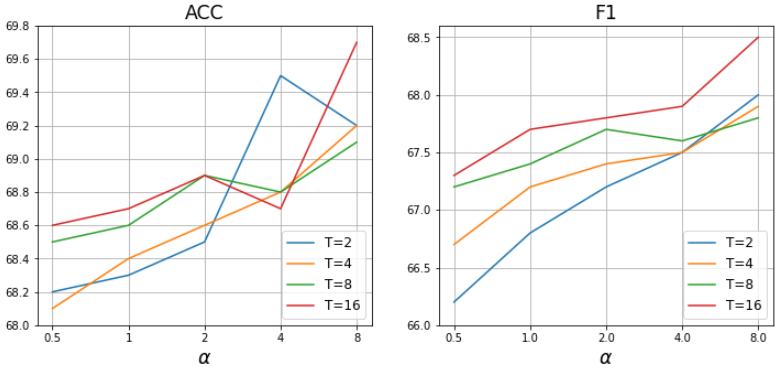} 
\caption{Ablation study on different Mixing parameter $\alpha$. $T$ is the number of samples generated from each original sample.} 
\label{fig:5} 
\end{figure}
\subsection{Ablation Study}

\textbf{Effects of data size:} In order to demonstrate the effect of Global Mixup in few sample scenarios, We conducted extended experiments on CNN, LSTM, and BERT. The results are shown in Table \ref{CNNfewtable}, \ref{few lstmtable} and \ref{few BERT}, the subset of the above data set was used for the experiments and  $N=\{500, 2000, 5000\}$. Since the all data of TREC dataset is $5452$, the experiment of TREC with $N=5000$ uses all data. We run experiments 10 times to report the mean and the standard deviation of accuracy (\%). When only 500 training samples are available, CNN improves accuracy by $7.2\% $ on YELP, LSTM improves accuracy by $7.3\%$ on YELP, and BERT improves accuracy by $7.7\%$ on SST-1. In addition, as the results in Table \ref{CNNfewtable} \{SST-1\} show, Global Mixup exceeds the effect of training 5000 samples without data augmentation by training only 500 samples. We also find that Global Mixup is more effective in the case of fewer samples.

\textbf{Effect of Different generation sample number $T$}: We also conducted experiments on the subset of yelp to show the performance impact of the number of samples generated per original sample. As shown in \textbf{Figure} \ref{fig:4}, Among the values $T=\{0, 2, 4, 8, 16, 20, 32, 64\}$. Compared to large datasets, small datasets are relatively sparsely distributed, so using Global Mixup on small datasets will achieve relatively more improvements. Also, based on the experimental results, we suggest that for any size dataset, the optimal range of $T$ should be between 4 and 20.

\textbf{Effects of different Mixing parameter $\alpha$}: We show the performance with different $\alpha$ in \textbf{Figure} \ref{fig:5}. The parameter $\alpha$ decides $\lambda \sim \operatorname{Beta} (\alpha, \alpha)$, the larger $\alpha$ will make $\lambda$ concentrate at 0.5, which means that the generated virtual samples are more likely to be further away from the original sample pairs. We choose $\alpha$ from $\{0.5, 1, 2, 4, 8\}$, we observed that $ \alpha =8$ achieved the best performance.

\section{Conclusion}
We propose Global Mixup, a new data augmentation method that transforms the previous one-stage augmentation process into two-stage, and solves the ambiguity problem caused by the linear interpolation of \textbf{Mixup} and \textbf{Mixup variants}. The experiment shows its superior performance, and its effect is more obvious when there are fewer samples. We believe if there is a better virtual examples generation strategy, Global Mixup will achieve better results, this is what we will explore in the future.



\bibliographystyle{kr}
\bibliography{kr-sample}

\end{document}